\documentclass[letterpaper, 10 pt, conference]{ieeeconf}  

\IEEEoverridecommandlockouts                              

\usepackage{balance}
\usepackage{graphicx}
\usepackage{xspace}
\usepackage{multirow}
\usepackage[table,xcdraw]{xcolor}
\usepackage{float}
\usepackage{amsmath} 
\usepackage{caption}
\usepackage{subcaption}
\usepackage{amsmath}

\usepackage{verbatim}

\newcommand{\system}{EscortE\xspace}
\newcommand{\reidsystem}{EscortE-ReID\xspace}
\newcommand{\actionsystem}{EscortE-Action\xspace}

\title{\LARGE \bf
Online Human Action Detection during Escorting
}

\author{Siddhartha Mondal$^{1}$, Avik Mitra$^{1}$, and Chayan Sarkar$^{1,2}$
\thanks{$^{1}$S. Mondal, A. Mitra, and C. Sarkar are with TCS Research, India. {\tt\small \{m.sidd9, mitra.avik1, sarkar.chayan\}@tcs.com}.}
\thanks{$^{2}$C. Sarkar is a Visiting Scholar at Yale University, United States. {\tt\small chayan.sarkar@yale.edu}}
}

\begin{document}

\maketitle
\thispagestyle{empty}
\pagestyle{empty}

\begin{abstract}
The deployment of robot assistants in large indoor spaces has seen significant growth, with escorting tasks becoming a key application. However, most current escorting robots primarily rely on navigation-focused strategies, assuming that the person being escorted will follow without issue. In crowded environments, this assumption often falls short, as individuals may struggle to keep pace, become obstructed, get distracted, or need to stop unexpectedly. As a result, conventional robotic systems are often unable to provide effective escorting services due to their limited understanding of human movement dynamics. To address these challenges, an effective escorting robot must continuously detect and interpret human actions during the escorting process and adjust its movement accordingly. However, there is currently no existing dataset designed specifically for human action detection in the context of escorting. Given that escorting often occurs in crowded environments, where other individuals may enter the robot’s camera view, the robot also needs to identify the specific human it is escorting (the subject) before predicting their actions. Since no existing model performs both person re-identification and action prediction in real-time, we propose a novel neural network architecture that can accomplish both tasks. This enables the robot to adjust its speed dynamically based on the escortee's movements and seamlessly resume escorting after any disruption. In comparative evaluations against strong baselines, our system demonstrates superior efficiency and effectiveness, showcasing its potential to significantly improve robotic escorting services in complex, real-world scenarios.
\end{abstract}
\section{INTRODUCTION}

Navigating through an airport can be particularly challenging for some people, especially when time is tight, which can raise anxiety levels. In such situations, having someone or something to take you to your destination, such as a boarding gate, can not only ease the stress of travel, but also help avoid delays in flight departure. This task is well-suited for robots. Beyond airports, robotic escorts have potential applications in shopping malls, hospitals, office premises, retail stores, and restricted facilities. For people with social anxiety, the ability to avoid unnecessary interactions with strangers when asking for directions can greatly reduce stress and discomfort. But the question remains: what does it take to transform a mobile robot into a functional escort service provider in real-world scenarios?

\begin{figure} [t]
    \centering
    \includegraphics[width=\linewidth]{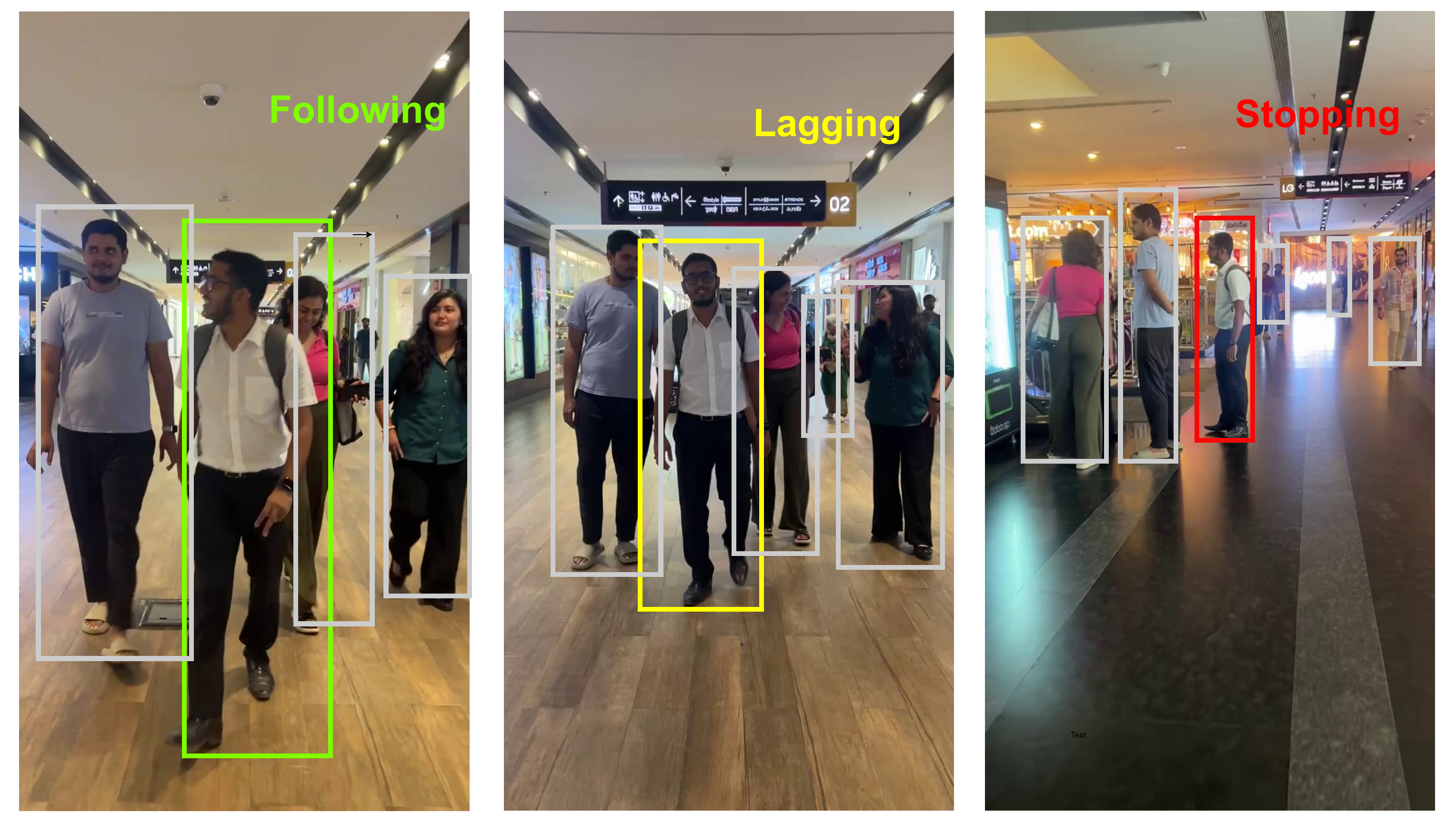}
    \caption{The three human action during escorting are -- following, lagging, and stopping. In a public place, multiple person can be visible along with the escortee. Hence, an escort robot needs to employ person re-identification and the three action detection to provide a better service.}
    \label{fig:data_sample}
\end{figure}

Current escort robots typically operate with little regard for the escortee's intentions and real-life movement dynamics. They navigate to the destination under the assumption that the user will follow them without issue~\cite{jiao2019path}. However, many escortees find it difficult to keep pace due to mobility limitations, crowded environments, or other distractions. Therefore, escortee-aware navigation is essential to deliver an effective and efficient user experience. In other words, the robot must be able to monitor the movements of the escortee and adapt its own motion accordingly while navigating to the desired destination. This requires a system for the detection of human actions, such as following, lagging, stopping, during escorting (Fig.~\ref{fig:data_sample}). In crowded environments, the robot must also employ a person re-identification system to handle situations where the escortee is obstructed by others or when multiple people are within the robot’s field of vision as shown in Fig.~\ref{fig:data_sample}. By using a re-identification mechanism, the robot can continue tracking the correct individual without confusion. This simple yet powerful solution has the potential to significantly enhance navigation experiences in large indoor environments, providing reliable assistance and guidance to those in need.

Traditionally, escorting robots use multiple sensors to track a person's movement during escorting. Ohya and Munekata~\cite{ohya2002intelligent} employed a wearable LED device, while Nomdedeu \textit{et al.}~\cite{nomdedeu2008experiment} proposed the use of wearable IMU devices for human tracking during escorting. More recently, Pérez \textit{et al.}~\cite{perez2023follow} developed a system based on the cognitive architecture of CORTEX, which integrates data from three different sensors: an RGBD camera, a face identification camera, and a microphone array. However, wearable sensors or multisensory solutions are not always practical for escorting applications in public spaces. Conte \textit{et al.}~\cite{conte2021autonomous} proposed an intent prediction model by the escorting robot based on head orientation and body pose. But it does not account for human speed variation or the presence of crowd in a public space.


In this work, we propose a vision-based approach to detect human action online so that the robot can be more aware of the escortee. Our primary contributions are twofold.

\begin{itemize}
    \item Due to the lack of suitable labeled datasets, we have created a new dataset of human motion applicable to various escorting scenarios, including crowded environments. This dataset will enable further research in this area.
    \item We present a novel neural network architecture that can perform both person re-identification and action prediction efficiently. This system operates in real-time and does not depend on wearables, depth cameras, or other sensors.
\end{itemize}

\section{RELATED WORK}
In this section, we briefly discuss the relevant works related to two key subsystems: person re-identification and human activity recognition.

\subsection{Person re-identification}
A comprehensive study by Leng \textit{et al.}~\cite{leng2019survey} provides an overview of existing methods in generalized re-identification (ReID), while Ye \textit{et al.}~\cite{ye2021deep} offers a more focused examination of deep learning approaches to tackle the re-ID problem. In surveillance applications, video-based person re-identification is typically performed across multiple cameras~\cite{mclaughlin2016recurrent, you2016top, li2018diversity}. Some approaches leverage graph convolution-based spatio-temporal methods~\cite{eom2021video, yang2020spatial}. To address the challenge of imperfect person detection due to changes in appearance, Gu \textit{et al.}~\cite{gu2020appearance} proposed a person representation method that adapts to such variations.

However, most of these studies focus on re-identification across multiple sources, aiming to re-identify a person across different images captured by non-overlapping cameras, predominantly in surveillance environments. This scenario typically involves images of the same person from various angles and different points of view. In contrast, the re-identification task in robotic escorting requires instantaneous, frame-wise recognition. Here, the robot must re-identify the subject as soon as it re-enters the field of view after being momentarily occluded or blocked by an obstacle, whether dynamic or static. This real-time requirement adds a unique challenge to the re-identification process in escorting applications.

\subsection{Human Activity Recognition}
Several studies provide comprehensive summaries of existing works on human activity recognition (HAR), highlighting various challenges and solution approaches~\cite{chen2021deep, islam2022human}. Mokhtari \textit{et al.}~\cite{mokhtari2022human} introduced a 3D skeleton data encoding method that generates images preserving the spatial and temporal dependencies between skeletal joints. In a subsequent study, they proposed an enhanced algorithm utilizing the concept of motion energy, which focuses on the most active skeleton joints during an action~\cite{10459847}. This encoding method improves discrimination between activities, leading to more accurate detection capabilities.

Online Action Detection (OAD) is a form of HAR where actions are detected in real time without access to the entire sequence of frames or videos. Chen \textit{et al.} proposed a transformer-based system called GATEHUB for OAD~\cite{chen2022gatehub}. More recently, another transformer-based system, MiniROAD, was introduced by An \textit{et al.}~\cite{an2023miniroad}, which outperforms existing models in terms of accuracy. However, due to the real-time nature of OAD, these models typically exhibit lower precision and accuracy compared to conventional HAR algorithms.

\section{System Overview}
\begin{figure}
    \centering
    \includegraphics[width=\linewidth]{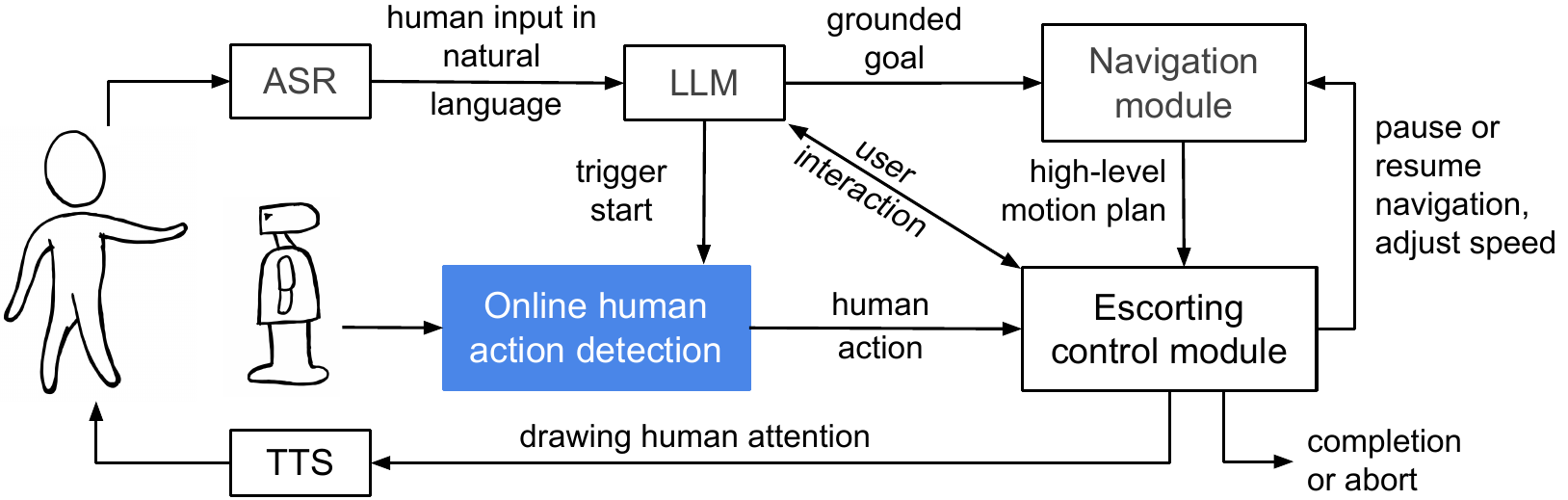}
    \caption{An overview of a robotic escorting system with online human action detection.}
    \label{fig:overview}
\end{figure}

\begin{figure*} []
    \centering    
    \includegraphics[width=0.98\linewidth]{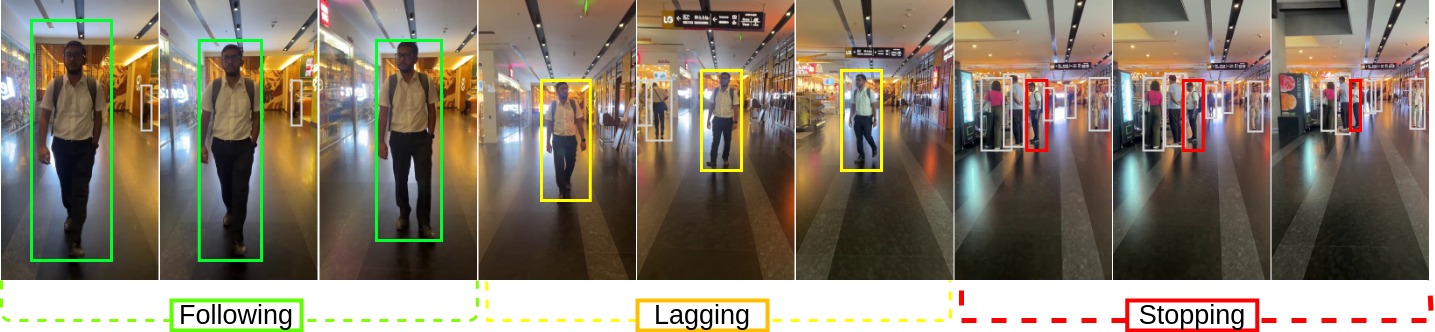}
    \caption{Sample frames of a sequence from \system dataset -- each frames of each sequence is annotated with action class, subject person bounding box, and the list of non-subject (negative) person bounding boxes. If the distance between the camera and the subject is less than 2 m (approximately), it is marked as following. If the distance is more than 2 m, it is marked as lagging. If the person becomes stationary, it is marked as stopping. }
    \label{fig:data_reel}
\end{figure*}

Developing an autonomous escorting robot presents multiple challenges. This section outlines a conceptual system architecture for a robotic escort system, as illustrated in Fig.~\ref{fig:overview}. The system comprises several key components: an automatic speech recognition (ASR) module, a natural language understanding and response generation module powered by a large language model (LLM), a navigation module, an online human action detection module, an intelligent escorting control module, and a text-to-speech (TTS) module. The goal of this system is to assist a human user in navigating from point A to point B within a crowded environment.

The escorting process begins with goal grounding, where the robot identifies the escortee’s intended destination. First, the spoken input from the escortee is converted into text using as ASR~\cite{pramanick2022visual}. Then, the textual input parsed using a natural language understanding module~\cite{pramanick2019enabling, sarkar2023tage}, which provides the desired location within the indoor environment. Once determined within the indoor environment, the robot must compute an optimal route while accounting for obstacles such as other humans and robots. If the user only asks for the route direction to a particular location, a verbal route description can be provided~\cite{pramanick2024much}. However, if the user expresses desire to be escorted, the robot can start navigating after prompting the user to follow it. The navigation algorithm must be dynamic and adaptable to changes in the environment. Furthermore, escortee-aware navigation necessitates real-time human action detection to monitor the escortee's movement patterns.

The system accounts for three primary scenarios -- (i)~Following: when the escortee moves consistently with the robot, (ii)~Lagging: when the escortee falls behind due to crowd obstructions or a slower walking pace, and (iii)~Stopping: when the escortee halts for various reasons, requiring the robot to adjust accordingly. To ensure smooth navigation and seamless re-planning, the robot must continuously monitor the escortee’s state, interact as necessary, and adjust its speed or stop when needed using the escort control module. Additionally, in crowded environments where multiple individuals are present, an effective re-identification mechanism is essential to maintain tracking accuracy.

\begin{table*}
\centering
\renewcommand{\arraystretch}{1.2}
\begin{tabular}{|l|r|p{1.8cm}|p{1cm}|p{2.3cm}|c|}
\hline
\rowcolor[HTML]{C0C0C0} 
\multicolumn{1}{|c|}{\cellcolor[HTML]{C0C0C0}\textbf{Dataset}} &
  \multicolumn{1}{c|}{\cellcolor[HTML]{C0C0C0}\textbf{\# identities}} &
  \textbf{\begin{tabular}[c]{@{}c@{}}Non-stationary \\ camera\end{tabular}} &
  \textbf{\begin{tabular}[c]{@{}c@{}}Multiple \\ cameras\end{tabular}} &
  \textbf{\begin{tabular}[c]{@{}c@{}}Detection \\ method\end{tabular}} &
  \textbf{\begin{tabular}[c]{@{}c@{}}Evaluation \\ metric\end{tabular}} \\ \hline
Market1501~\cite{zheng2015scalable} & 1501 & No & Yes & DPM  & mAP \\ \hline
\rowcolor[HTML]{EFEFEF} 
RAiD~\cite{Das2014}  & 43   & No & Yes & Hand & CMC \\ \hline
CUHK03~\cite{li2014deepreid} & 1360 & No & Yes & DPM  & CMC \\ \hline
\rowcolor[HTML]{EFEFEF} 
VIPeR~\cite{richter2017playing}  & 632  & No & Yes & hand & CMC \\ \hline
i-LIDS~\cite{li2018unsupervised} & 119  & No & Yes & hand & CMC \\ \hline
\rowcolor[HTML]{EFEFEF} 
CUHK01~\cite{li2012human}     & 971  & No & Yes & hand & CMC \\ \hline
CUHK02~\cite{li2013locally}     & 1816 & No & Yes & hand & CMC \\ \hline
\rowcolor[HTML]{EFEFEF} 
CAVIAR~\cite{cheng2011custom} & 72   & No & Yes & hand & CMC \\ \hline
\textbf{\reidsystem} & \textbf{29} & \textbf{Yes} & \textbf{No} & \textbf{\begin{tabular}[l]{@{}l@{}}YOLOS with \\manual supervision\end{tabular}} &
\textbf{mAP} \\ \hline
\end{tabular}%
\caption{Summary of \reidsystem and other existing datasets for person re-identification.}
    \label{tab:reid_sota_data}
\end{table*}

\begin{table*}[]
\centering
\renewcommand{\arraystretch}{1.2}
\begin{tabular}{|c|c|c|c|c|c|c|c|c|c|}
\hline
\rowcolor[HTML]{C0C0C0} 
\textbf{Dataset} &
  \textbf{\begin{tabular}[c]{@{}c@{}}\#\\ Seq(s)\end{tabular}} &
  \textbf{\begin{tabular}[c]{@{}c@{}}Total \\ hours\end{tabular}} &
  \textbf{\begin{tabular}[c]{@{}c@{}}\# \\ subjects\end{tabular}} &
  \textbf{\begin{tabular}[c]{@{}c@{}}\# \\ actions\end{tabular}} &
  \textbf{\begin{tabular}[c]{@{}c@{}}Atomic \\ actions\end{tabular}} &
  \textbf{RGB} &
  \textbf{RGBD} &
  \textbf{Pose/Key-point} &
  \textbf{\begin{tabular}[c]{@{}c@{}}Other \\ sensors\end{tabular}} \\ \hline
OAD~\cite{li2016online}    & 59   & \_\_ & \_\_ & 10 & No  & Yes & Yes & Yes & No  \\ \hline
\rowcolor[HTML]{EFEFEF} 
PAMAP2~\cite{misc_pamap2_physical_activity_monitoring_231} & \_\_ & \_\_ & 9    & 18 & No  & No  & No  & No  & Yes \\ \hline
HAR~\cite{anguita2013public}    & \_\_ & \_\_ & 30   & 6  & No  & No  & No  & No  & Yes \\ \hline
\rowcolor[HTML]{EFEFEF} 
MMFiT~\cite{stromback2020mm}  & \_\_ & \_\_ & \_\_ & 10 & No  & No  & Yes & Yes & Yes \\ \hline
DAHLIA~\cite{vaquette2017daily} & 51   & 33.4 & 44   & 8  & Yes & Yes & Yes & Yes & No  \\ \hline
\rowcolor[HTML]{EFEFEF} 
MoVi~\cite{ghorbani2020movi}   & \_\_ & 32.6 & 90   & 21 & No  & Yes & No  & Yes & Yes \\ \hline
HOMAGE~\cite{rai2021home} & 5700 & 30   & \_\_ & 70 & Yes & Yes & No  & No  & Yes \\ \hline
\rowcolor[HTML]{EFEFEF} 
\textbf{\actionsystem} & \textbf{359} & \textbf{1.42} & \textbf{29} & \textbf{3} & \textbf{No} & \textbf{Yes} & \textbf{No} & \textbf{No} & \textbf{No} \\ \hline
\end{tabular}
%
\caption{Summary of \actionsystem and other existing dataset for (offline and online) human action detection.}
\label{tab:actionDetection_sota_data}
\end{table*}

A key feature of this architecture is its real-time human action detection module, which integrates re-identification and action prediction. When a user approaches, the system captures an initial reference frame to facilitate consistent tracking. The robot then records a video stream of the user and extracts embeddings from the initial reference frame and subsequent frames using a vision transformer (ViT)~\cite{dosovitskiy2020image}. These embeddings are compared against all detected individuals in each frame to identify and track the subject throughout the escorting process.

Once the escortee is identified, the system analyzes a series of frames to classify the escortee’s movement into one of the three predefined states:
\begin{itemize}
    \item Following: The relative distance between the robot and escortee remains stable.
    \item Lagging: The distance gradually increases over several frames, prompting the robot to slow down or notify the escortee to keep up.
    \item Stopping: The escortee remains stationary, leading the robot to halt and prompt the escortee to proceed. If the escortee does not respond, the robot decides whether to continue or terminate the escorting process.
\end{itemize}
By integrating these components, the proposed system enables an escorting robot to navigate effectively in dynamic, crowded environments while maintaining robust human-robot interaction and adaptive behavior.


\section{Dataset Description}

As there is no publicly available dataset featuring a robot escorting a person in a public place, we created a new dataset, \system. This dataset is annotated for two key tasks: person re-identification and online human action detection. The corresponding sub-systems for these tasks are referred to as \reidsystem and \actionsystem, respectively. The dataset can be used for either task with the appropriate annotations.

\system consists of videos capturing individuals walking in crowded environments. It comprises mobile camera recordings (720p resolution at 30 fps) featuring 29 subjects. To ensure diversity and reduce bias, we carefully selected subjects with variations in physique, age, complexion, and gender. Additionally, we incorporated a range of lighting conditions, crowd densities, face and body visibility levels, and distances from the camera to enhance dataset variability.

Table~\ref{tab:reid_sota_data} presents a comparison of existing person re-identification datasets against \system, while Table~\ref{tab:actionDetection_sota_data} compares \actionsystem with publicly available datasets in Human Activity Recognition (HAR) and Online Action Detection (OAD).

\subsection{Data collection}
Data collection took place in a public indoor setting (a shopping mall), with multiple subjects participating each day. Each participant took turns being the subject in various scenarios. In each scenario, the designated escort subject was instructed to follow the camera, simulating the behavior of following an escort robot. Meanwhile, other participants walked randomly to create a realistic public space environment. Additionally, the presence of random pedestrians often led to occlusions of the subject, further enhancing the dataset’s complexity.

The dataset includes three distinct actions (Fig.~\ref{fig:data_reel}).
\begin{itemize}
    \item Following: The subject followed the camera while maintaining an approximate distance of less than 2 meters.
    \item Lagging: The subject intentionally slowed down while the camera operator maintained a constant speed, increasing the distance to more than 2 meters. The movement was carefully adjusted to be perceived as lagging by human observers.
    \item Stopping: The subject remained stationary at a specific point.
\end{itemize}

For each subject, multiple short video sequences (approximately 12 per subject) were recorded. Each sequence contained either a single action, a combination of two actions, all three actions, or repeated actions in a random order. Additionally, the sequences varied in the number of non-subject participants appearing in the frame and the frequency of subject occlusions. Subjects were instructed to perform the actions naturally, according to their own judgment.

To ensure accurate data collection, approximate distance measurements were taken using floor tiles as reference points, and a second person was present to guide the subjects in initiating movement and maintaining the required distances. To further validate manual distance annotations, a standard depth estimation system~\cite{yang2024depth} was applied to the video sequences. All participants consented to the use of their videos for research purposes and agreed to make them publicly available.

\subsection{Data annotation}
Each frame of every video sequence was annotated to ensure accurate labeling. First, we used the YOLOS object detector~\cite{fang2021you} to identify all person instances and generate corresponding bounding boxes. Next, we manually labeled the detected individuals as either subject or non-subject. If YOLOS failed to detect a person, we manually added a bounding box, and any incorrect detections were removed. Then, frame-wise action annotations were performed, allowing action recognition to be implemented either sequentially or on a per-frame basis. The dataset consists of a total of 359 sequences from 29 subjects, which are split into 250 training sequences, 49 development sequences, and 60 test sequences.

\section{Network Architecture}
\begin{figure}
    \centering
    \includegraphics[width=\linewidth]{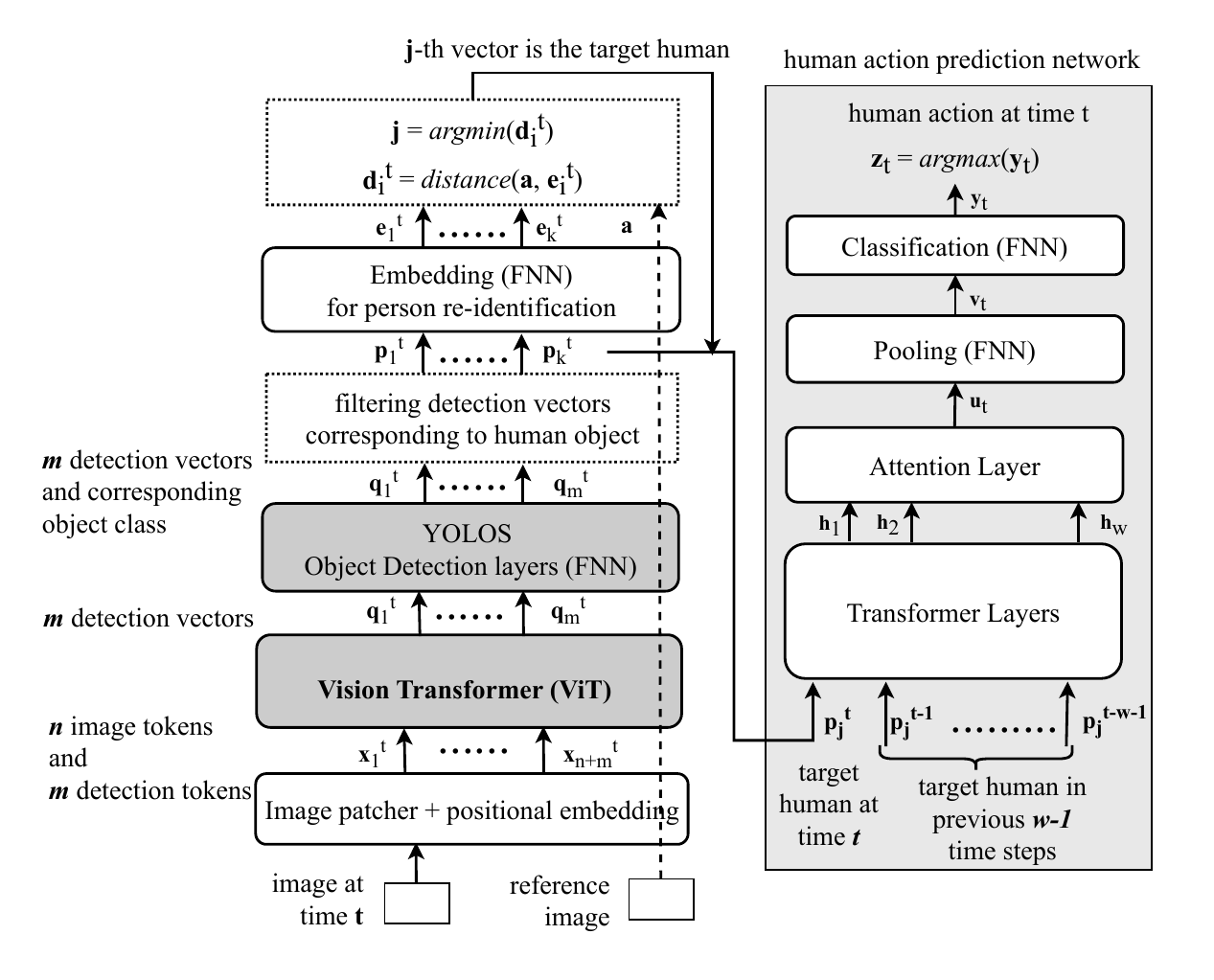}
    \caption{Network architecture for online human action detection along with person re-identification.}
    \label{fig:neural_network}
\end{figure}

The online human action detection module performs two key sub-tasks: person re-identification and action detection. Traditionally, these tasks have been addressed separately in existing literature. To achieve real-time inference, we propose an architecture that integrates both tasks (Fig.~\ref{fig:neural_network}).

Our network is built on the YOLOS object detector~\cite{fang2021you}, which employs a Vision Transformer (ViT)~\cite{dosovitskiy2020image} as its backbone. Given an input image, YOLOS first divides it into a set of image patches, which are then transformed into tokens. These tokens, along with detection tokens (initialized as zero vectors), are processed by the ViT. The output of ViT consists of an equal number of vectors corresponding to the input tokens (both image and detection tokens). YOLOS then classifies the detection token vectors into known object classes or background. Our network subsequently filters out only those detection vectors classified as persons.

For person re-identification, the filtered person vectors are passed through an embedding layer composed of two linear layers with a ReLU activation in between. A reference embedding vector is also computed from a reference image using the same process. To determine identity, we compute the L2 norm between the reference vector and all person vectors in the input image. The person vector with the smallest distance below a predefined threshold is identified as the subject. If no person vector meets this criterion, the subject is considered absent from the image. Empirically, we decided the threshold value to be 1.5.

To train the embedding layer, we utilize the \textbf{triplet margin loss} function~\cite{hermans2017defense, zeng2020hierarchical}. 
\begin{gather}
L(a,p,n) = \max\{d(a_i, p_i) - d(a_i, n_i) + \text{margin}, 0\} \\
\textit{where}, \;\;\;\; d(x_i, y_i)=\|x_i - y_i\|_2 \nonumber
\end{gather}
The triplet loss measures relative similarity by comparing positive ($p_i$) and negative ($n_i$) vectors with respect to the anchor (reference) vector ($a_i$). This approach enhances the network's ability to distinguish between individuals effectively.

For action detection, the subject person vectors from each input image are stored in a buffer of window size $w$, corresponding to the same number of input frames. These vectors are then passed through four transformer layers, producing $w$ output vectors. These vectors are subsequently processed by a self-attention layer to compute a single vector ($u_t$) representing the entire window of images. The vector $u_t$ of dimension $d$ is obtained as a weighted average of the $w$ vectors:
\begin{eqnarray}
B_{w\times1} & = & H_{w\times d} * A_{d\times1} \nonumber \\
C_{w\times1} & = & softmax(B_{w\times1}) \nonumber \\
u_{d\times1} & = & H^T_{d\times w} * C_{w\times1}, \nonumber
\end{eqnarray}
where $H_{w\times d}$ represents the output vectors of the transformer. The vector $u_t$ is then passed through a pooling layer with ReLU activation before being fed into a classification layer. The weights of the transformer, attention, pooling, and classification layers are learned using the \textbf{cross-entropy loss}~\cite{zhang2018generalized}.
\begin{eqnarray}
L(y, p) = - (y \log(p) + (1 - y) \log(1 - p)),
\end{eqnarray}
where $p$ is the prediction logit and $y$ is the target.

Both components of the network are trained separately using their respective loss functions. During action detection training, ground truth person identification is utilized to ensure accurate learning.

\section{Result}
Robotic escorting is a real-world application that requires real-time inference of human action detection module while being independent of any extra sensors or wearable on the subject. Hence, the baselines for comparison were chosen accordingly. In the following subsections, we discuss the evaluation metric, comparison of person re-identification and action detection with their corresponding state-of-the-art, and evaluation of the end-to-end \system module with three strong baselines.

\begin{table}[t]
\centering
\renewcommand{\arraystretch}{1.2}
\begin{tabular}{|l|l|}
\hline
\rowcolor[HTML]{C0C0C0} 
\textbf{Model}                & \textbf{Precision} \\ \hline
\rowcolor[HTML]{EFEFEF} 
Face recognition using DeepFace & 23.14              \\ \hline
Face recognition using VGGFace  & 62.59              \\ \hline
\rowcolor[HTML]{EFEFEF} 
CLiP-ReID           & 65.88              \\ \hline
\reidsystem             & 90.44              \\ \hline
\end{tabular}%
\caption{Performance comparison of \reidsystem and the state-of-the-art models for person re-identification on \system dataset.}
\label{tab:reid_comparison}
\end{table}

\begin{table}[t]
\centering
\renewcommand{\arraystretch}{1.2}
\begin{tabular}{|l|l|}
\hline
\rowcolor[HTML]{C0C0C0} 
\textbf{Model}         & \textbf{Precision} \\ \hline
\rowcolor[HTML]{EFEFEF} 
ESTIE (Human activity recognition)     & 44.99              \\ \hline
miniROAD (Online action detection) & 59.30              \\ \hline
\rowcolor[HTML]{EFEFEF} 
\actionsystem     & 79.40              \\ \hline
\end{tabular}%
\caption{Performance comparison of \actionsystem and state-of-the-art human activity recognition and online action detection model on \system dataset.}
    \label{tab:action_comparison}
\end{table}

\begin{figure}[t]
    \centering
    \begin{subfigure}[b]{0.48\linewidth}
        \centering        \includegraphics[width=\textwidth]{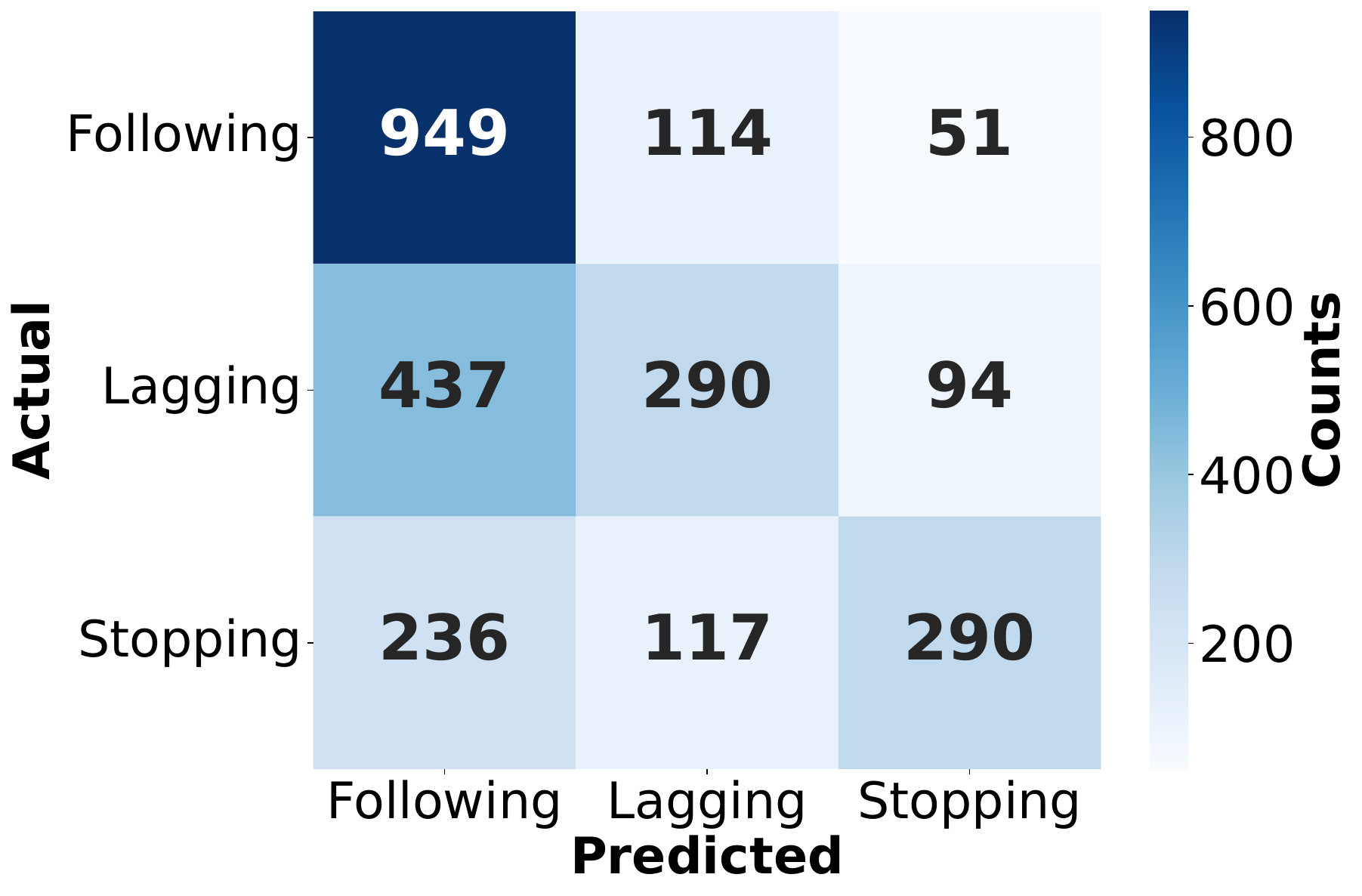}
        \caption{miniROAD}
        \label{fig:conf_matrix_miniroad}
    \end{subfigure}
    \hfill
    \begin{subfigure}[b]{0.48\linewidth}
        \centering
        \includegraphics[width=\textwidth]{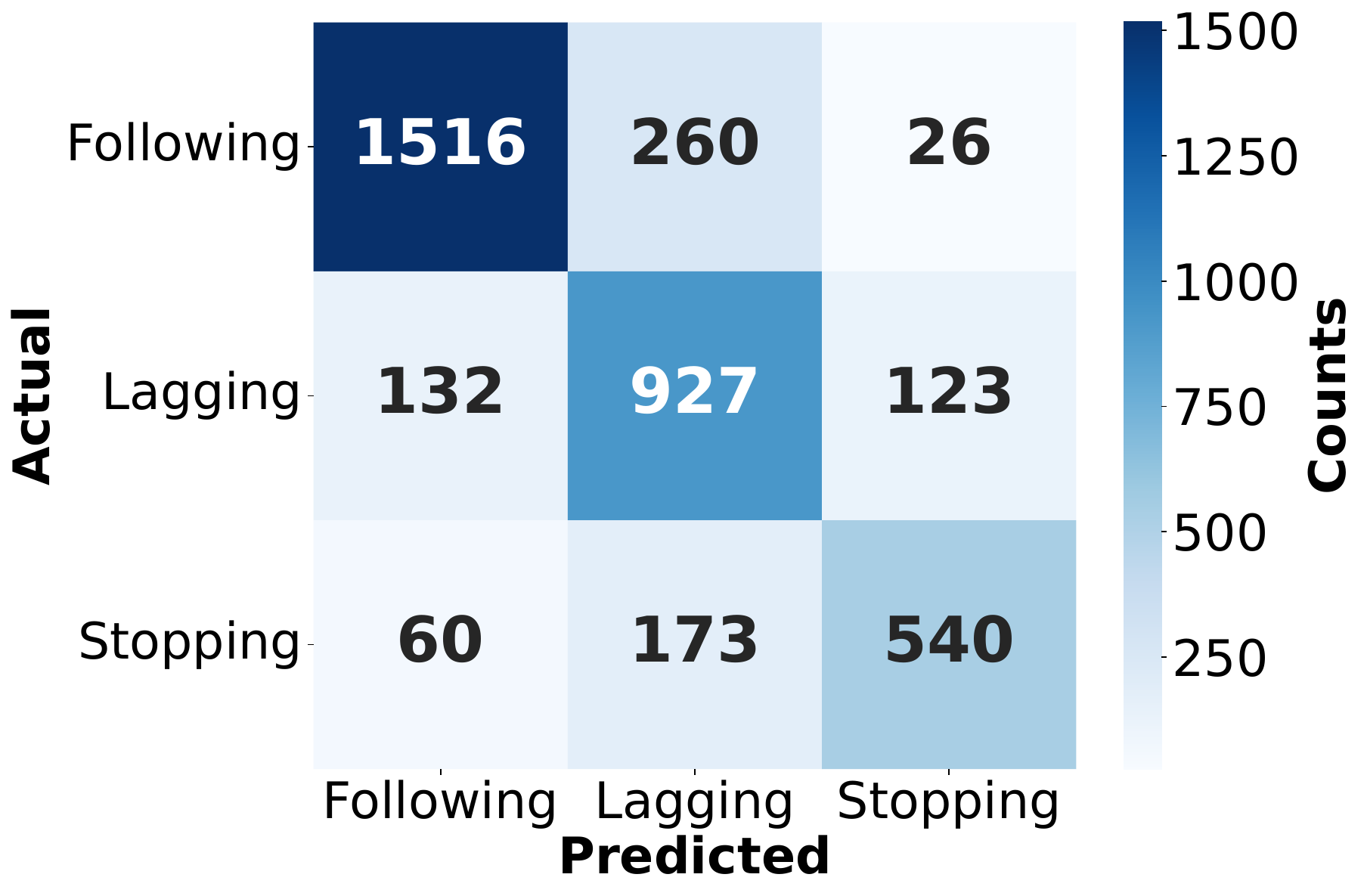}
        \caption{\actionsystem}
        \label{fig:conf_matrix_escorte_action}
    \end{subfigure}
    \caption{Comparison of confusion matrices for action prediction between miniROAD and \actionsystem.}
    \label{fig:conf_escorteAction}
\end{figure}

\begin{table*}[t]
\centering
\renewcommand{\arraystretch}{1.2}
\begin{tabular}{|l|l|}
\hline
\rowcolor[HTML]{C0C0C0} 
\textbf{System} & \textbf{Pipeline composition of the systems} \\ \hline
\rowcolor[HTML]{EFEFEF} 
Baseline1 &
  \begin{tabular}[c]{@{}l@{}}Faster R-CNN (object detection) + Imagenet (pre-trained) + CLiP ReID + Resnet50 (feature extractor) + miniROAD (fine-tuned)\end{tabular} \\ \hline
Baseline2          & \begin{tabular}[c]{@{}l@{}}Faster R-CNN (object detection) + Imagenet (pre-trained) + CLiP ReID  + ViT embedder + \actionsystem \end{tabular} \\ \hline
\rowcolor[HTML]{EFEFEF} 
Baseline3 &
  \begin{tabular}[c]{@{}l@{}}ViT embedder + YOLOS object detector + \reidsystem + Resnet50 (feature extractor) + miniROAD (fine-tuned) \end{tabular} \\ \hline
Proposed           & \begin{tabular}[c]{@{}l@{}}ViT embedder + YOLOS object detector + \reidsystem + \actionsystem \end{tabular}                          \\ \hline
\end{tabular}%
    \caption{Composition of several strong baselines for person re-identification and online human action detection together along with our proposed architecture as shown in Fig.~\ref{fig:neural_network}.}
    \label{tab:baseline_def}
    \end{table*}
    
\begin{table}[t]
\centering
\renewcommand{\arraystretch}{1.2}
\begin{tabular}{|l|l|}
\hline
\rowcolor[HTML]{C0C0C0} 
\textbf{System}                  & \textbf{Precision} \\ \hline
\rowcolor[HTML]{EFEFEF} 
Baseline1: CLiP ReID + miniROAD      & 17.75              \\ \hline
Baseline2: CLiP ReID + \actionsystem & 24.85              \\ \hline
\rowcolor[HTML]{EFEFEF} 
Baseline3: \reidsystem + miniROAD    & 40.88              \\ \hline
\system: \reidsystem + \actionsystem & 78.09              \\ \hline
\end{tabular}%
\caption{Performance comparison among the baseline systems as defined in Table~\ref{tab:baseline_def} and our proposed system.}
    \label{tab:joint_comparison}
\end{table}

\begin{figure}[t]
    \centering
    \includegraphics[width=0.95\linewidth]{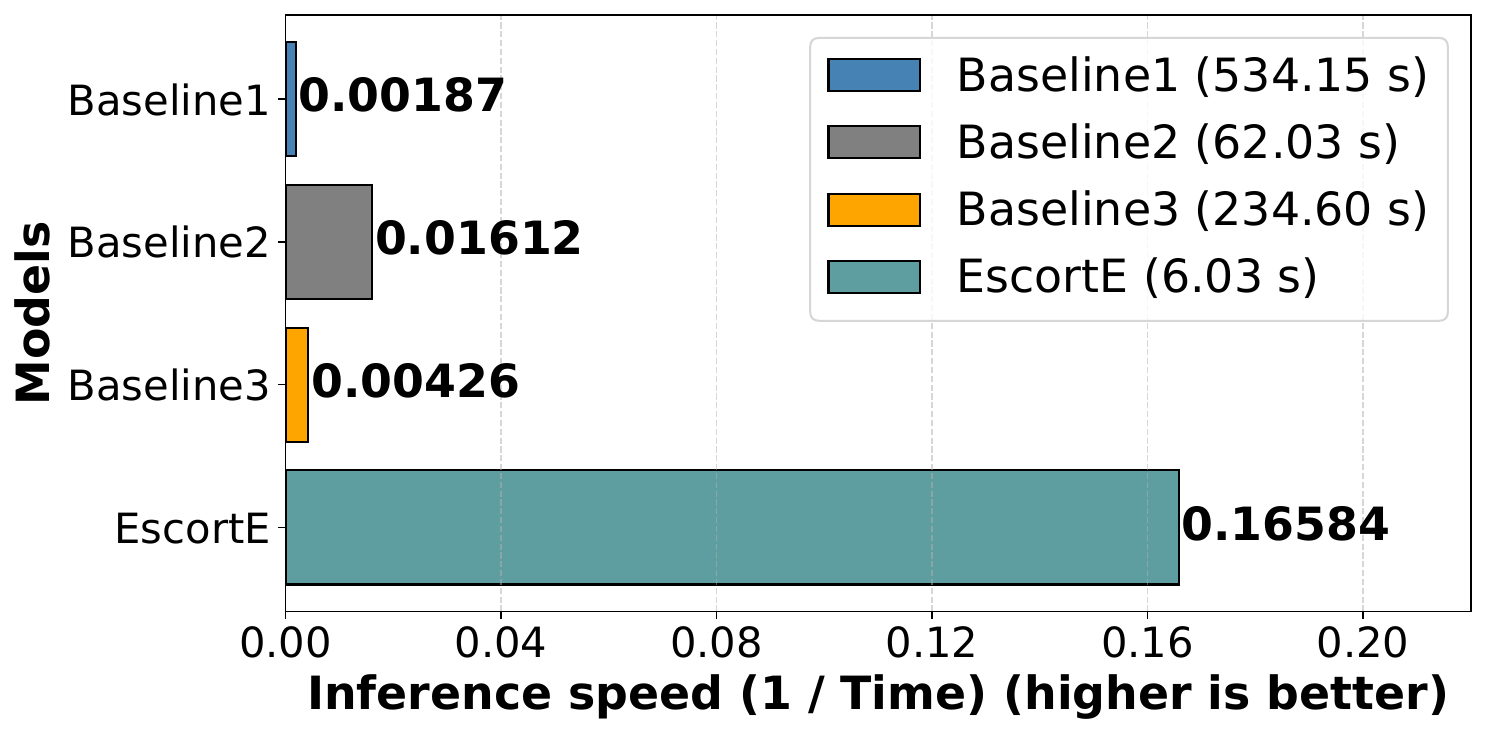}
    \caption{\system inference time comparison with baseline systems.}
    \label{fig:inf_speed}
\end{figure} 

\subsection{Evaluation Metric}
We use mean average precision (mAP) to evaluate both sub-tasks individually and the joint system. mAP is a widely used metric, calculated by averaging the average precision (AP) across all classes. In addition, we calculate the inference time of the joint system to assess its real-time performance. The inference time is calculated using the following formula:
\begin{equation}
    \mathrm{t}_{i} = \alpha \cdot [(w-1) \cdot \mathrm{t}_{r}] - \mathrm{t}_{f} + \mathrm{t}_{r} + \mathrm{t}_{a}, 
    \quad \alpha =
    \begin{cases} 
      0 & \text{if } \mathrm{t}_{r} > \mathrm{t}_{f} \\
      1 & \text{if } \mathrm{t}_{r} \leq \mathrm{t}_{f}
    \end{cases}
    \label{eq:inference_time}
\end{equation}
where \( t_{i} \) is the inference time, \( w \) is the window length, \( t_{r} \) is the re-identification time per frame, \( t_{f} = 1/fps\) is time gap between two frames, and \( t_{a} \) is the action recognition time for the window of frames. In other words, the inference time effectively calculates the time required to perform the re-identification for all the frames in a window and the action detection, minus the capture time of the frames. 

For all our experiments, we used 2\;s window data both for testing and evaluation. Since the frames are recoded at 30 fps, this leads to $w=60$.

\subsection{Re-identification evaluation}
We assess our re-identification system with respect to the state-of-the-art models of face recognition and person re-identification. Table~\ref{tab:reid_comparison} presents an overview of \reidsystem against SOTA in terms of precision. The fall in the precision of face recognition models \cite{taigman2014deepface} for person re-identification tasks is mainly because there are multiple frames with occluded faces and partially visible faces, which is common in escorting system. Additionally, there are many frames where the subject is considerably far from the camera. Hence, the extraction of facial features becomes an ordeal. To address these challenges, Li \textit{et al.} proposed CLiP-ReID~\cite{li2023clip} which performs better compared to face recognition-based models. However, \reidsystem has higher precision compared to both models. 

\subsection{Action detection evaluation}
To assess how our proposed network performs in terms of action detection in the context of escorting, we compared it with two state-of-the-art models -- (i)~ESTIE~\cite{mokhtari2022human}, which specializes in human activity recognition(HAR) and (ii)~miniROAD~\cite{an2023miniroad}, which performs best for online action detection (OAD) task. Both models were fine-tuned with the \system dataset under the same conditions as \actionsystem. Table~\ref{tab:action_comparison} summarizes the performance of the models. ESTIE did not perform up to the mark for the action classes relevant to escorting, whereas miniROAD showed much better results compared to ESTIE. However, \actionsystem outperforms both models. We use both miniROAD and \actionsystem to build baselines in the joint setup.

To further evaluate the performance of miniROAD and \actionsystem in class-wise prediction, we present the confusion matrices in Fig.~\ref{fig:conf_escorteAction}. These matrices clearly show that both models perform well in predicting the ``following'' class, with occasional confusion between ``following'' and ``lagging''. However, when it comes to predicting the ``lagging'' and ``stopping'' classes, miniROAD exhibits more frequent misclassifications compared to our system.

\subsection{Joint system evaluation}
Next, we have compared \system with a joint setup against three baseline systems defined in Table~\ref{tab:baseline_def}. The best-performing model for re-identification and action detection was chosen and combined to build different joint systems for a better understanding of their performances if they are integrated to work with a full-fledged escort system. The systems often demanded preprocessing or pretrained models, which have also been added while building the baselines.   

We have evaluated the precision of each system against all the baselines defined in Table~\ref{tab:baseline_def}. The results indicated in Table~\ref{tab:joint_comparison} show that our proposed system outperforms all other baseline systems in precision by good margins. 

In addition to this, we have also calculated the inference time for each baseline and our proposed system using Eq.~\ref{eq:inference_time}. As observed in Fig.~\ref{fig:inf_speed}, the inference time of \system compared to other baselines is significantly faster and it is 97.49\% faster than its closest rival.

\section{CONCLUSION}
Robotic escort systems are invaluable for guiding individuals through large, unfamiliar indoor spaces. However, existing systems often overlook escortee awareness, assuming the subject will simply follow the robot. In reality, the escortee may struggle to keep up in crowded environments, get distracted, or even stop. To address this, robots must adjust their speed or stop based on the escortee’s actions. The three key actions to detect during escorting are: following, lagging, and stopping. Additionally, in crowded settings, the robot may momentarily lose sight of the escortee due to occlusion, or other people might enter its field of view. Therefore, continuous person re-identification is crucial. The absence of relevant public datasets led us to create a custom one. This paper introduces the Human Escorting Dataset (named \system), with annotations for both person re-identification (\reidsystem) and action detection (\actionsystem). We also propose a novel neural network architecture that jointly handles both tasks, unlike existing systems that treat them separately. We demonstrate the efficacy of our sub-systems through comparisons with state-of-the-art models and show that our joint system outperforms strong baselines with significantly higher precision as well as lower inference time.

In the future, we aim to enhance the accuracy of the action detection subsystem by incorporating distance information, which can be estimated from 2D images. Additionally, we plan to build the complete escorting system as outlined in Fig.~\ref{fig:overview}.


\balance



\bibliographystyle{IEEEtran}
\bibliography{main}

\end{document}